\documentclass[conference]{IEEEtran}
\IEEEoverridecommandlockouts
\usepackage{cite}
\usepackage{amsmath,amssymb,amsfonts}
\usepackage{algorithmic}
\usepackage{graphicx}
\usepackage{textcomp}
\usepackage{xcolor}
\usepackage{amsmath}
\usepackage{subcaption}
\usepackage{url}
\def\BibTeX{{\rm B\kern-.05em{\sc i\kern-.025em b}\kern-.08em
    T\kern-.1667em\lower.7ex\hbox{E}\kern-.125emX}}
\begin{document}

\title{Predicting Parkinson's Disease using Latent Information extracted from Deep Neural Networks\\
}

\author{\IEEEauthorblockN{Ilianna Kollia}
\IEEEauthorblockA{\textit{Big Data \& Analytics Center} \\
\textit{IBM Hellas}\\
Athens, Greece \\
ikollia@gr.ibm.com}
\and
\IEEEauthorblockN{Andreas-Georgios Stafylopatis}
\IEEEauthorblockA{\textit{School of Electrical \& Computer Engineering} \\
\textit{National Technical University of Athens}\\
Athens, Greece \\
andreas@cs.ntua.gr}
\and
\IEEEauthorblockN{Stefanos Kollias}
\IEEEauthorblockA{\textit{School of Computer Science}\\
\textit{University of Lincoln}\\
Lincoln, United Kingdom \\
skollias@lincoln.ac.uk}
}


\maketitle

\begin{abstract}
 This paper presents a new method for medical diagnosis of neurodegenerative diseases, such as Parkinson's, by extracting and using latent information from trained Deep convolutional, or convolutional-recurrent Neural Networks (DNNs). In particular, our approach adopts a combination of transfer learning, \textit{k}-means clustering and \textit{k}-Nearest Neighbour classification of deep neural network learned representations to provide enriched prediction of the disease based on MRI and/or DaT Scan data. A new loss function is introduced and used in the training of the DNNs, so as to perform adaptation of the generated learned representations between data from different medical environments. Results are presented using a recently published database of Parkinson's related information, which was generated and evaluated in a hospital environment.   
\end{abstract}


\section{Introduction}
Machine learning techniques have been largely used in medical signal and image analysis for prediction of neurodegenerative disorders, such as Alzheimer's and Parkinson's, which significantly affect elderly people, especially in developed countries~\cite{d1},~\cite{d2},~\cite{d3}.

In the last few years, the development of deep learning technologies has boosted the investigation of using deep neural networks for early prediction of the above-mentioned neurodegenerative disorders. In~\cite{d4}, stacked auto-encoders have been used for diagnosis of Alzheimer's disease.{3-D~Convolutional} Neural Networks (CNNs) have been used in~\cite{d5} to analyze imaging data for Alzheimer's diagnosis. Both methods were based on the Alzheimer's disease neuroimaging initiative dataset, including medical images and assessments of several hundred subjects. Recently, CNNs and convolutional-recurrent neural network (CNN-RNN) architectures have been developed for prediction of Parkinson's disease~\cite{new2}, based on a new database including Magnetic Resonance Imaging (MRI) data and Dopamine Transporters (DaT) Scans from patients with Parkinson's and non patients~\cite{new3}.

In this paper we focus on the early prediction of Parkinson's. It is the above two types of medical image data, i.e. MRI and DaT Scans that we explore for predicting an asymptomatic (healthy) status, or the stage of Parkinson's at which a subject appears to be. In particular, MRI data show the internal structure of the brain, using magnetic fields and radio waves. An atrophy of the Lentiform and Caudate Nucleus can be detected in MRI data of patients with Parkinson's. DaT Scans are a specific form of single-photon emission computed tomography, using Ioflupane Iodide-123 to detect lack of dopamine in patients' brain.   

In the paper we base our developments on the deep neural network (DNN) structures (CNNs, CNN-RNNs) developed in~\cite{new2} for predicting Parkinson's using MRI, or DaT Scan, or MRI \& DaT Scan data from the recently developed Parkinson's database~\cite{new3}. We extend these developments by extracting latent variable information from the DNNs trained with  MRI \& DaT Scan data and generate clusters of this information; these are evaluated by medical experts with reference to the corresponding status/stage of Parkinson's. The generated and medically annotated cluster centroids are used next in three different scenarios of major medical significance:

1) Transparently predicting a new subject's status/stage of Parkinson's; this is performed using nearest neighbor classification of new subjects' MRI and DaT scan data with reference to the cluster centroids and the respective medical annotations. 

2) Retraining the DNNs with the new subjects' data, without forgetting the current medical cluster annotations; this is performed by considering the retraining as a constrained optimization problem and using a gradient projection training algorithm instead of the usual gradient descent method.

3) Transferring the learning achieved by DNNs fed with MRI \& DaT scan data, to medical centers that only possess MRI information about subjects, thus improving their prediction capabilities; this is performed through a domain adaptation methodology, in which a new error criterion is introduced that includes the above-derived cluster centroids as desired outputs during training.   

Section II describes related work where machine learning techniques have been applied to MRI and DaT Scan data for detecting Parkinson's. The new Parkinson's database we are using in this paper is also described in this section. 
Section III  first describes the extraction of latent variable information from trained deep neural networks and then presents the proposed approach in the framework of the three considered testing, transfer learning and domain adaptation scenarios. Section IV provides the experimental evaluation which illustrates the performance of the proposed approach using an augmented version of the Parkinson's database, which we also make publicly available. Conclusions and future work are presented in Section V.

\section{Related Work}\label{rel_work}

Medical image data constitute a rich source of information regarding cell degeneration in the human nervous system of Parkinson's patients. MRI and DaT Scan data have been the focus of related research; in~\cite{d10}, principal component analysis and support vector machines were applied to MRI data, while the same techniques and empirical mode decomposition were applied to DaT Scans in~\cite{d11}.    

A Parkinson's database comprising MRI and DaT Scan data from 78 subjects, 55 patients with Parkinson's and 23 non patients, has been recently released~\cite{new3}; it includes, in total 41528 MRI data (31147 from patients and 10381 from non patients) and 925 DaT scans (595 and 330 respectively). Our developments next are based on this database.   

CNN architectures~\cite{ylc86},~\cite{c10} include convolutional, pooling and fully connected layers, in which convolutional kernel and fully connected layer weights are usually learned through gradient descent, while pooling layers reduce the input sizes through averaging operations. CNN-RNN architectures~\cite{c10},~\cite{d12} are capable of effectively analyzing temporal variations of the inputs, by permitting intra layer connections and using appropriate gating operations. 

Recent advances in deep neural networks~\cite{c10},~\cite{new},~\cite{c9},~\cite{b5}
have been explored in~\cite{new2}, where convolutional (CNN) and convolutional-recurrent (CNN-RNN) neural networks were developed and trained to classify the information in the above Parkinson's database in two categories, i.e., patients and non patients, based on either MRI inputs, or DaT Scan inputs, or together MRI and DaT Scan inputs. 

DaT Scans, which are a specific examination for Parkinson's, generally convey more information than MRI; however, using both inputs can provide better prediction performances.   

\begin{figure}[htbp]

\begin{center}
\includegraphics[scale=0.25]{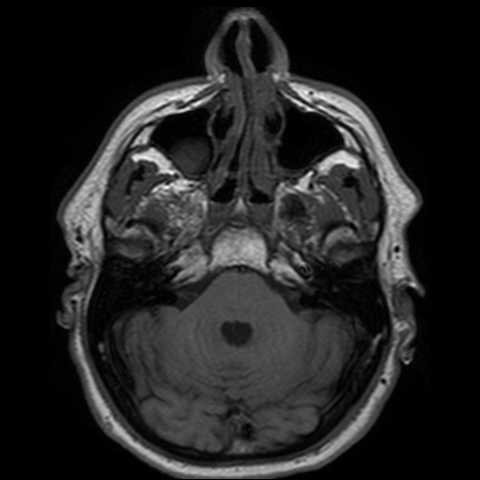}
\includegraphics[scale=0.25]{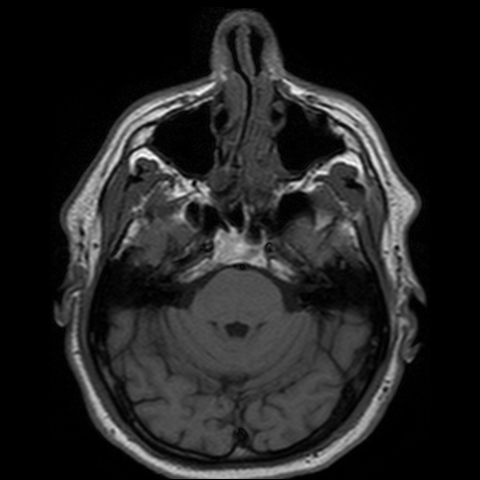}
\includegraphics[scale=0.255]{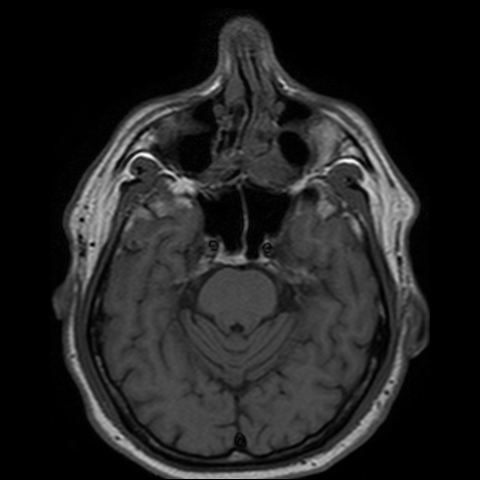}
\includegraphics[scale=0.27]{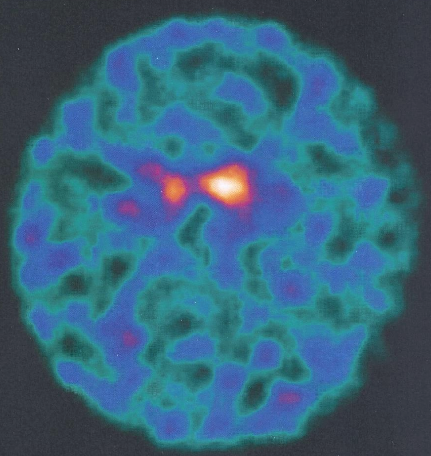}
\end{center}
\caption{An MRI triplet of consecutive frames and a corresponding DaT Scan}
\label{fig}
\end{figure}

The developed networks included: transfer learning of the ResNet-50 network~\cite{c25} as far as the convolutional part of the networks was concerned, with retraining of the fully connected network layers; adding on top of this and training a recurrent network using Gated Recurrent Units (GRU)~\cite{c6} in an end-to-end manner.

In this paper we focus first on the analysis of the combined MRI and DaT Scan dataset. It should be mentioned that the target in Parkinson's disease detection through MRI data is on estimation of the volume of the lentiform and of the capita of the caudate nucleus. To deal with volume estimation, we analyse MRIs in triplets of consecutive frames. 
Thus, an MRI triplet of (gray-scale) images and a DaT Scan (colour) image constitute the input to the CNN and/or CNN-RNN architectures that we use in our developments. Fig.~\ref{fig} shows such a triplet of consecutive frames from an MRI sequence and a corresponding DaT Scan image. 

Section III.A presents the methodology used to extract latent variables from the trained DNNs and to achieve diagnosis of Parkinson's. Section III.B describes the approach for retraining the DNNs with new information, while preserving the already extracted information. In Section III.C we examine DNN-based analysis of only MRI input triplets and show how this analysis can be improved by adaptation of the latent variable information extracted from the DNNs trained with both MRI and DaT Scan data.    

\section{The Proposed Approach}

\subsection{\textbf{Extracting Latent Variables from Trained Deep Neural Networks}}

The proposed approach begins with training a CNN, or a CNN-RNN architecture, on the (train) dataset of MRI and DaT Scan data. The CNN networks include a convolutional part and one or more Fully Connected (FC) layers, using neurons with a ReLU activation function. In the CNN-RNN case, these are followed by a recurrent part, including one ore more hidden layers, composed of GRU neurons. 

We then focus on the neuron outputs in the last FC layer (CNN case), or in the last RNN hidden layer (CNN-RNN case). These latent variables, extracted from the trained DNNs, represent the higher level information through which the networks produce their predictions, i.e., whether the input information indicates that the subject is patient, or not.  

In particular, let us consider the following dataset for training the DNN to predict Parkinson's:
\begin{equation}
\label{eq: training dataset}
\mathcal{P} = \big\{(\textbf{x}(j), d(j)); \ j=1,\ldots,n\big\} 
\end{equation}
and the corresponding test dataset:
\begin{equation}
\label{eq: test dataset}
\mathcal{Q} = \big\{(\widetilde{\textbf{x}}(j), \widetilde{d}(j)); \ j=1,\ldots,m\big\} 
\end{equation}
where: $\textbf{x}(j)$ and $d(j)$ represent the $n$ network training inputs (each of which consists of an MRI triplet and a DaT Scan) and respective desired outputs (with a binary value 0/1, where 0 represents a non patient and 1 represents a patient case);  $\widetilde{\textbf{x}}(j)$ and $\widetilde{d}(j)$ similarly represent the $m$ network test inputs and respective desired outputs.

After training the Deep Neural Network using dataset $\mathcal{P}$, its $l$ neurons' outputs in the final FC, or hidden layer,   $\{\textbf{r}(j)\}$ and $\{ \widetilde{\textbf{r}}(j)\}$, both $\in\mathbb{R}^{l}$, are extracted as latent variables, obtained through forward-propagation of each image, in train set $\mathcal{R}_p$ and test set $\mathcal{R}_q$ respectively:
\begin{equation}
\label{eq:traininglatent}
\mathcal{R}_p = \big\{(\textbf{r}(j), \ j=1,\ldots,n\big\} 
\end{equation}
and 
\begin{equation}
\label{eq:testlatent}
\mathcal{R}_q = \big\{(\widetilde{\textbf{r}}(j),  \ j=1,\ldots,m\big\} 
\end{equation}

The following clustering procedure is then implemented on the $\{\textbf{r}(j)\}$ in $\mathcal{R}_p$:

We generate a set of clusters ${T} =\{\textbf{t}_1,\ldots,\textbf{t}_k\}$ by minimizing the within-cluster $L^{2}$ norms of the function 
\begin{equation}
\label{eq:kmeans}
\widehat{{T}}_{k\text{-means}} = \underset{{T}}{\operatorname{arg\ min}} 
\sum_{j=1}^{k} \sum_{\mathbf{r}\in {R}_{p}}^{} 
\big|\big|\textbf{r}-\textbf{$\mu$}_{j}\big|\big|^{2}
\end{equation}
where $\textbf{$\mu$}_{j}$ is the mean value of the data in cluster $j$.

This is done using the k-means++~\cite{d50} algorithm, with the first cluster centroid $\textbf{u}(1)$ being selected at random from ${T}$. The class label of a given cluster is simply the mode class of the data points within it.

As a consequence, we generate a set of cluster centroids, representing the different types of input data included in our train set $\mathcal{P}$:
\begin{equation}
\label{eq: cluster centroid set}
\mathcal{U} = \big\{(\textbf{u}(j), \ j=1,\ldots,k\big\} 
\end{equation}

Through medical evaluation of the MRI and DaT Scan images corresponding to the cluster centroids, we can annotate each cluster according to the stage of Parkinson's that its centroid represents. 

By computing the euclidean distances between the test data in $\mathcal{R}_q$ and the cluster centroids in $\mathcal{U}$ and by then using the nearest neighbor criterion, we can assign each one of test data to a specific cluster and evaluate the obtained classification - disease prediction - performance. This is an alternative way to the prediction accomplished when the trained DNN is applied to the test data. 

This alternative prediction is, however, of great significance: in the case of non-annotated new subject's data,  selecting the nearest cluster centroid in $\mathcal{U}$ can be a transparent way for diagnosis of the subject's Parkinson's stage; the available MRI and DaT Scan data and related medical annotations of the cluster centroids being compared to the new subject's data.

\subsection{\textbf{Retraining of Deep Neural Networks with Annotated Latent Variables}}
\label{subsec-Adaptation}

Whenever new data, either from patients, or from non patients, are collected, they should be used to extend the knowledge already acquired by the DNN, by adapting its weights to the new data. In such a case, let us assume that a new train dataset, say $\mathcal{P}_1$, usually of small size, say $s$, is generated and an updated DNN should be created based on this dataset as well.   

There are different methods developed in the framework of transfer learning~\cite{d20}, for training a new DNN on $\mathcal{P}_1$ using the structure and weights of the above-described DNN. However, a major problem is that  of catastrophic forgetting, i.e., the fact that the DNN forgets some formerly learned information when fine-tuning to the new data. This can lead to loss of annotations related to the latent variables extracted from the formerly trained DNN. To avoid this, we propose the following DNN adaptation method, which preserves annotated latent variables. 

For simplicity of presentation, let us consider a CNN architecture, in which we keep the convolutional and pooling layers fixed and retrain the FC and output layers.  
Let $\textbf{W}$ be a vector including the weights of the FC and output network layers of the original network, before retraining, and $\textbf{W}'$ denote the new (updated) weight vector, obtained through retraining. Let us also denote by, $\textbf{w}$ and $\textbf{w}'$, respectively, the weights connecting the outputs of the last FC, defined as $\textbf{r}$ in Eq.~(\ref{eq:traininglatent}), to the network outputs, $y$.

During retraining, the new network
weights, $\textbf{W}'$, are computed by minimizing the following error criterion:
\begin{equation}\label{eq1}
    \mathcal{E} = \mathcal{E}_{\mathcal{P}_1} + \lambda \cdot \mathcal{E}_{\mathcal{P}}
\end{equation}
where $\mathcal{E}_{\mathcal{P}_1}$ represents the misclassifications done in $\mathcal{P}_1$, which includes the new data and $\mathcal{E}_{\mathcal{P}}$ represents the misclassifications in $\mathcal{P}$, which includes the old information. $\lambda$ is used to differentiate the focus between the new and old data. In the following we make the hypothesis that a small change of the weights $\textbf{W}$ is enough to achieve good classification performance in the current conditions. Consequently, we get:
\begin{equation}\label{eq2}
    \textbf{W}' = \textbf{W} + \Delta \textbf{W}
\end{equation}
and in the output layer case:

\begin{equation}\label{eq3}
    \textbf{w}' = \textbf{w} + \Delta \textbf{w}
\end{equation}
in which $\Delta \textbf{W}$ and $\Delta \textbf{w}$ denote small weight increments. Under this formulation, we can apply a first-order Taylor series expansion to make neurons' activation linear.  

Let us now give more attention to the new data in $\mathcal{P}_1$. We can do this, by expressing $\mathcal{E}_{\mathcal{P}_1}$ in Eq.~(\ref{eq1}) in terms of the following constraint:

\begin{equation}\label{eq4}
    y'(j) = d(j); \ j=1,\ldots,s
\end{equation}
which requests that the new network outputs and the desired outputs are identical. 

Moreover, to preserve the formerly extracted latent variables, we move the input data corresponding to the annotated cluster centroids in $\mathcal{U}$ from dataset $\mathcal{P}$ to $\mathcal{P}_1$. Consequently, Eq.~(\ref{eq4}) includes these inputs as well; the size of $\mathcal{P}_1$ becomes:

\begin{equation}\label{eq24}
    s' = s+ k
\end{equation}
where $k$ is the number of clusters in $\mathcal{U}$.

Let the difference of the retrained network output  $y'$ from the original one  $y$ be:

\begin{equation}\label{eq5}
    \Delta y(j) = y'(j) - y(j)
\end{equation}

Expressing the output $y'$ as a
weighted average of the last FC layer outputs $\textbf{r}'$ with the $\textbf{w}'$ weights, we get~\cite{new2}

\begin{equation}\label{eq6}
    y'(j) = y(j) + f^{h} \cdot \textbf{w} \cdot \Delta \textbf{r}(j) + \Delta \textbf{w} \cdot \textbf{r}(j)
\end{equation}
where $f^h$ denotes the derivative of the former DNN output layer's neurons' activation function.
Inserting Eq.~(\ref{eq4}) into Eq.~(\ref{eq6}) results in:

\begin{equation}\label{eq7}
    d(j) - y(j) = f^{h} \cdot \textbf{w} \cdot \Delta \textbf{r}(j) + \Delta \textbf{w} \cdot \textbf{r}(j)
\end{equation}

All terms in Eq.~(\ref{eq7}) are known, except of the differences in weights $\Delta \textbf{w}$ and last FC neuron outputs $\Delta \textbf{r}$. As a consequence, Eq.~(\ref{eq7}) can be used to compute the new DNN weights of the output layer in terms of the neuron outputs of the last FC layer.

If there are more than one FC layers, we apply the same procedure, i.e., linearize the difference of the $\textbf{r}'$, iteratively through the previous FC layers and express the $\Delta \textbf{r}$ in terms of the weight differences in these layers. When reaching the  convolutional/pooling layers, where no retraining is to be performed, the procedure ends, since the respective $\Delta \textbf{r}$ is zero. It can be shown, similarly to~\cite{new2} that the weight updates $\Delta \textbf{W}$ are finally estimated through the solution of a set of linear equations defined on $\mathcal{P}_1$ :

\begin{equation}\label{eq8}
    \textbf{v} = V \cdot \Delta \textbf{W}
\end{equation}
where matrix $V$ includes weights of the original DNN and vector $\textbf{v}$ is defined as follows:

\begin{equation}\label{eq9}
    \text{v}(j) = d(j) - y(j); \ j=1,\ldots,s'
\end{equation}
with $y(j)$ denoting the output of the original DNN applied to the data in $\mathcal{P}_1$.

Similarly to~\cite{new2}, the size of $\textbf{v}$ is lower than the size of $\Delta \textbf{W}$; many methods exist, therefore, for solving  Eq.~(\ref{eq9}). Following the assumption made in the beginning of this section, we choose the solution that provides minimal modification of the original DNN weight. This is the one that provides the minimum change in the value of $\mathcal{E}$ in 
Eq.~(\ref{eq1}). 

Summarizing, the targeted adaptation can be solved as a nonlinear constrained optimization problem, minimizing Eq.~(\ref{eq1}), subject to Eq.~(\ref{eq4}) and the selection of minimal weight increments. In our implementation, we use the gradient projection method~\cite{c40} for computing the network weight updates and consequently the adapted DNN architecture.  

\subsection{\textbf{Domain Adaptation of Deep Neural Networks through Annotated Latent Variables}}

In the two previous subsections we have focused on generation, based on extraction of latent variables from a trained DNN, and use of cluster centroids for prediction and adaptation of a Parkinson's diagnosis system. To do this, we have considered all available imaging information, consisting of MRI and DaT Scan data. 

However, in many cases, especially in general purpose medical centers, DaT Scan equipment may not be available, whilst having access to MRI technology. In the following we present a domain adaptation methodology, using the annotated latent variables extracted from the originally trained DNN, to improve prediction of Parkinson's achieved when using only MRI input data. A new DNN training loss function is used to achieve this target.  

Let us consider the following train and test datasets, similar to $\mathcal{P}$ and $\mathcal{Q}$ in Eq.~(\ref{eq: training dataset}) and  Eq.~(\ref{eq: test dataset}) respectively, in which the input consists only of triplets of MRI data:

\begin{equation}
\label{eq: train1}
\mathcal{P}' = \big\{(\textbf{x}'(j), d'(j)); \ j=1,\ldots,n'\big\}  
\end{equation}
and
\begin{equation}
\label{eq: test1}
\mathcal{Q}' = \big\{(\widetilde{\textbf{x}'}(j), \widetilde{d}'(j)); \ j=1,\ldots,m'\big\} 
\end{equation}
where: $\textbf{x}'(j)$ and $d'(j)$ represent the $n'$ network
training inputs (each of which consists of only an MRI triplet) and respective desired outputs (with a binary value $0/1$, where 0 represents a non patient and 1 represents a patient case);  $\widetilde{\textbf{x}'}(j)$ and $\widetilde{d}'(j)$ similarly represent the $m'$ network test inputs and respective desired outputs.

Using $\mathcal{P}'$, we train a similar DNN structure - as in the full MRI and DaT Scan case -, producing the following vector of $l$ neuron outputs in its last FC or hidden layer:

\begin{equation}
\label{eq:train1latent}
\mathcal{R}'_p = \big\{(\textbf{r}'(j), \ j=1,\ldots,n'\big\} 
\end{equation}
with the dimension of each $\textbf{r}'$ vector being $l$, as in the original DNN last FC, or hidden, layer.

A far as the $\textbf{r}'$ outputs are concerned, it would be desirable that these latent variables being closer, e.g., according to the mean squared error criterion, to one of the cluster centroids in Eq.~(\ref{eq: cluster centroid set}) that belongs to the same category(patient/non patient) with them.
 
 In this way, training of the DNN with only MRI inputs, would also bring its output $y'$ closer to the one generated by the original DNN; this would potentially improve the network's performance, towards the much better one produced by the original DNN (trained with both MRI and DaT Scan data). 

Let us compute the euclidean distances between the latent variables in $\mathcal{R}'_p$ and the cluster centroids in $\mathcal{U}$ as defined in Eq.~(\ref{eq: cluster centroid set}). Using the nearest neighbor criterion we can define a set of desired vector values for the $\textbf{r}'$ latent variables, with respect to the $k$ cluster centroids, as follows:

\begin{equation}
\label{eq:clusterlatent}
\mathcal{Z}_p = \big\{(z(i,j),  i=1,\ldots,k;  j=1,\ldots,n'\big\} 
\end{equation}
where $z(i,j)$ is equal to, either 1 in the case of the cluster centroid $\textbf{u}(i)$ that was selected, as closest to $\textbf{r}'(j)$ during the above-described procedure, or equal to 0 in the case of the rest cluster centroids.

In the following, we introduce the $z(i,j)$ values in a modified Error Criterion to be used in DNN learning to correctly classify the MRI inputs.

Normally, the DNN (CNN, or CNN-RNN) training is performed through minimization of the error criterion in Eq.~(\ref{eq:5}) in terms of the DNN weights:

\begin{equation} \label{eq:5}
\mathcal{E}_1 = \frac{1}{n'} \sum_{j=1}^{n'}  (d'(j)-y'(j))^2 
\end{equation}

where $y'$ and $d'$ denote the actual and desired network outputs and $n'$ is equal to the number of all MRI input triplets.

We propose a modified Error Criterion, introducing an additional term, using the following definitions:

\begin{equation} \label{eq:6}
\textbf{g}(i,j) = \textbf{u}(i) - \textbf{r}'(j), \ i=1,\ldots,k; \ j=1,\ldots,n'
\end{equation}
and
\begin{equation} \label{eq:7}
G(i,j) = \textbf{g}(i,j)* (\textbf{g}(i,j))^T
\end{equation}
with \textit{T} indicating the transpose operator.

It is desirable that the $G(i,j)$ term - with a respective value of $z(i,j)$ equal to one - is minimized, whilst the $G(i,j)$ values - corresponding to the rest of the $z(i,j)$ values, which are equal to zero - are maximized. Similarly to~\cite{c8}, we pass $G(i,j)$ through a softmax \textit{f} function and subtract its output from 1, so as to obtain the above-described respective minimum and maximum values.  

The generated Loss Function is expressed in terms of the differences of the transformed $G(i,j)$ values from the corresponding desired responses $z(i,j)$, as follows:
\begin{equation} \label{eq:8}
\mathcal{E}_2 = \frac{1}{kn'}  \sum_{i=1}^{k} \sum_{j=1}^{n'} (z(i,j) - [1-f(G(i,j)])^2  
\end{equation}
calculated on the $n'$ data and the $k$ cluster centroids.

In general, our target is to minimize together Eq.~(\ref{eq:5}) and Eq.~(\ref{eq:8}). We can achieve this using the following Loss Function:
\begin{equation} \label{eq:9}
\mathcal{E}_{new} = \eta \mathcal{E}_1+ (1-\eta) \mathcal{E}_2                               
\end{equation}
where $\eta$ is chosen in the interval [$0$, $1$]. 

Using a value of $\eta$ towards zero provides more importance to the introduced centroids of the clusters of the latent variables extracted from the best performing DNN, trained with both MRI and DaT Scan data. On the contrary, using a value towards one leads to normal error criterion minimization.

\section{Experimental Evaluation}\label{experimentalSection}
In this section we present a variety of experiments for evaluating the proposed approach. The implementation of all algorithms described in the former Section has been performed in Python using the Tensorflow library.

\subsection{The Parkinson's Dataset}
The data that are used in our experiments come from the Parkinson's database described in Section~\ref{rel_work}. For training the CNN and CNN-RNN networks, we performed an augmentation procedure in the train dataset, as follows. After forming all triplets of consecutive MRI frames, we generated combinations of these image triplets with each one of the DaT Scans in each category (patients, non patients). 

Consequently, we created a dataset of 66,176 training inputs, each of them consisting of 3 MRI and 1 DaT Scan images. In the test dataset, which referred to different subjects than the train dataset, we made this combination per subject; this created 1130 test inputs.      

For possible reproduction of our experiments, both the training and test datasets, each being split in two folders - patients and non patients - are available upon request from the \url{mlearn.lincoln.ac.uk} web site.

\subsection{Testing the proposed Approach for Parkinson's Prediction}

We used the DNN structures described in~\cite{new2}, including both CNN and CNN-RNN architectures to perform Parkinson's diagnosis, using the train and test data of the above-described database. The convolutional and pooling part of the architectures was based on the ResNet-50 structure; GRU units were used in the RNN part of the CNN-RNN architecture.

The best performing CNN and CNN-RNN structures, when trained with both MRI and DaT Scan data, are presented in       Table~\ref{table:results1}.

\begin{table*}[!t]
\caption{DNN best performing structures on DaT Scan and MRI data}
\label{table:results1}
\centering

\setlength{\arrayrulewidth}{.4pt}
\begin{tabular}{|c|c|c|c|c|c|}
\hline \hline
Structure & No FC layers & No Hidden Layers & No Units in FC Layer(s) & No Units in Hidden Layers & Accuracy ($\%$) \\
\hline
CNN & 2 & - & 2622-1500 & - & {94}\% \\
CNN-RNN & 1 & 2 & 1500 & 128-128 & 98\% \\
\hline \hline
\end{tabular}
\end{table*}

It is evident that the CNN-RNN architecture was able to provide excellent prediction results on the database test set. We, therefore, focus on this architecture for extracting latent variables. For comparison purposes, it can be mentioned that the performance of a similar CNN-RNN architecture when trained only with MRI inputs was about 70\%. 

It can be seen, from Table~\ref{table:results1}, that the number $l$ of neurons in the last FC layer of the CNN-RNN architecture was 128. This is, therefore, the dimension of the vectors $\textbf{r}$ extracted as in Eq.~(\ref{eq:traininglatent}) and used in the cluster generation procedure of
Eq.~(\ref{eq:kmeans}). 

We then implemented this cluster generation procedure, as described in the former Section. The k-means algorithm provided five clusters
of the data in the 128-dimensional space.  Fig. 2 depicts a 3-D visualization of the five cluster centroids; stars in blue color denote the two centroids corresponding to non patient data, while squares in red color represent the three cluster centroids corresponding to patient data.    

\begin{figure}[htbp]
\begin{center}
\includegraphics[scale=0.7]{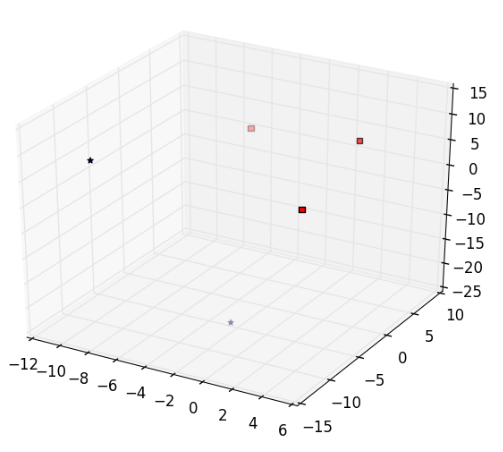}
\end{center}
\caption{The five cluster centroids in 3-D; 2 of them (stars, blue color) depict non patients and 3 of them (squares, red color) represent patients}
\label{fig2}
\end{figure}

\begin{figure*}
\centering
\begin{subfigure}{.3\textwidth}
  \centering
  \includegraphics[scale=0.8]{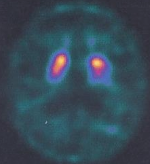}
  \caption{Centroid $\textbf{t}_1$}
\end{subfigure}%
\begin{subfigure}{.3\textwidth}
  \centering
  \includegraphics[scale=0.8]{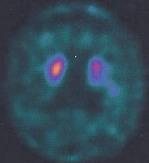}
  \caption{Centroid $\textbf{t}_2$}
\end{subfigure}\\
\centering
\begin{subfigure}{.3\textwidth}
  \centering
  \includegraphics[scale=1.53]{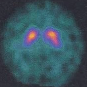}
  \caption{Centroid $\textbf{t}_3$}
\end{subfigure}
\begin{subfigure}{.3\textwidth}
   \centering
   \includegraphics[scale=0.6]{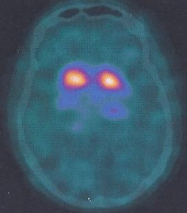}
  \caption{Centroid $\textbf{t}_4$}
\end{subfigure}
\begin{subfigure}{.3\textwidth}
   \centering
   \includegraphics[scale=0.47]{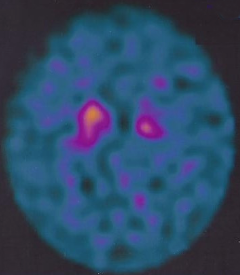}
  \caption{Centroid $\textbf{t}_5$}
\end{subfigure}
\caption{The DaT Scans corresponding to the five cluster centroids  $\textbf{t}_1$ - $\textbf{t}_5$ }
\label{fig3}
\end{figure*}

With the aid of medical experts, we generated annotations of the images (3 MRI and 1 DaT Scan) corresponding to the 5 cluster centroids. It was very interesting to discover that these centroids represent different levels of Parkinson's evolution. Since the DaT Scans conveyed the major part of this discrimination, we show in Fig.3 the DaT Scans corresponding to each one of the cluster centroids. 

According to the provided medical annotation, the 1st centroid ($\textbf{t}_1$) corresponds to a typical non patient case. The 2nd centroid ($\textbf{t}_2$) represents a non patient case as well, but with some findings that seem to be pathological. Moving to the patient cases, the 3rd centroid ($\textbf{t}_3$) shows an early step of Parkinson's - in stage 1 to stage 2, while the 4th centroid ($\textbf{t}_4$) denotes a typical Parkinson's case - in stage 2. Finally, the 5th centroid ($\textbf{t}_5$) represents an advanced step of Parkinson's - in stage 3.

It is interesting to note here that, although the  DNN was trained to classify input data in two categories - patients and non patients -, by extracting and clustering the latent variables, we were able to generate a richer representation of the diagnosis problem in five categories. It should be mentioned that the purity of each generated cluster was almost perfect. 
\begin{table}[!t]
\caption{Training data in each generated cluster}
\label{table:results2}
\centering

\setlength{\arrayrulewidth}{.4pt}
\begin{tabular}{|c|c|}
\hline \hline
Cluster & No of Data (\%)  \\
\hline
$\textbf{t}_1$ & 4,3 \\ $\textbf{t}_2$ & 38,4 \\ $\textbf{t}_3$ & 27,6 \\ $\textbf{t}_4$  & 2,3 \\ $\textbf{t}_5$ & 27,4 \\
\hline \hline
\end{tabular}
\end{table}

Table~\ref{table:results2} shows the percentages of training data included in each one of the five generated clusters. It should be mentioned that almost two thirds of the data belong in clusters 2 and 3, i.e., in the categories which are close to the borderline between patients and non patients. These cases require major attention by the medical experts and the proposed procedure can be very helpful for diagnosis of such subjects' cases.

We tested this procedure on the Parkinson's test dataset, by computing the euclidean distances of the corresponding extracted latent variables from the 5 cluster centroids and by classifying them to the closest centroid. 

Table~\ref{table:results3} shows the number of test data referring to six different subjects that were classified to each cluster. All non patient cases were correctly classified. In the patient cases, the great majority of the data of each patient were correctly classified to one of the respective centroids. In the small number of misclassifications, the disease symptoms were not so evident. However, based on the large majority of correct classifications, the subject would certainly attract the necessary interest from the medical expert.

\begin{table}[!t]
\caption{Classification of 6 subjects' data in clusters $\textbf{t}_1$-$\textbf{t}_5$}
\label{table:results3}
\centering
\setlength{\arrayrulewidth}{.4pt}
\begin{tabular}{|c|c|c|c|c|c|}
\hline \hline
Test case & $\textbf{t}_1$ & $\textbf{t}_2$ & $\textbf{t}_3$ & $\textbf{t}_4$ & $\textbf{t}_5$   \\
\hline
Non Patient 1 & 44 & \textbf{398} & 0 & 0 & 0 \\ 
Non Patient 2 & 10 & \textbf{90} & 0 & 0 & 0 \\
Patient 1 & 3 & 7 & \textbf{94} & 8 & 8 \\ 
Patient 2  & 1 & 7 & \textbf{139} & 17 & 20 \\ 
Patient 3 & 3 & 0 & \textbf{145} & 18 & 38 \\ 
Patient 4 & 0 & 0 & 0 & 8 & \textbf{72} \\
\hline \hline
\end{tabular}
\end{table}

We next examined the ability of the above-described DNN to be retrained using the procedure described in Subsection~III.B. 

In the developed scenario, we split the above test data in two parts: we included 3 of them (Non Patient 2, Patient 2 and Patient 3) in the retraining dataset $\mathcal{P}_1$ and let the other 3 subjects in the new test dataset. The size $s'$ of $\mathcal{P}_1$ was equal to 493 inputs, including the five inputs corresponding to cluster centroids in $\mathcal{U}$; the size of the new test set was equal to 642 inputs.

We applied the proposed procedure to minimize the error over all train data in $\mathcal{P}$ and $\mathcal{P}_1$, focusing more on the latter, as described by Eq.~(\ref{eq4}). 

The network managed to learn and correctly classify all 493 $\mathcal{P}_1$ inputs, including the inputs corresponding to the cluster centroids, with a minimal degradation of its performance over $\mathcal{P}$ input data. We then applied the trained network to the test dataset consisting of three subjects. In this case, there was also a slight improvement, since the performance was raised to 98,91\%, compared to the corresponding performance on the same three subjects' data, shown in Table III, which was 98,44\%.  

Table~\ref{table:results4} shows the clusters to which the new extracted latent variables $\widetilde{\textbf{r}}$ were classified. A comparison with the corresponding results in~Table~\ref{table:results3} shows the differences produced through retraining.

\begin{table}[!t]
\caption{Classification of 3 subjects' data, after retraining, in clusters $\textbf{t}_1$-$\textbf{t}_5$}
\label{table:results4}
\centering
\setlength{\arrayrulewidth}{.4pt}
\begin{tabular}{|c|c|c|c|c|c|}
\hline \hline
Test case & $\textbf{t}_1$ & $\textbf{t}_2$ & $\textbf{t}_3$ & $\textbf{t}_4$ & $\textbf{t}_5$   \\
\hline
Non Patient 1 & 41 & \textbf{401} & 0 & 0 & 0 \\ 
Patient 1 & 2 & 5 & \textbf{99} & 7 & 7 \\ 
Patient 4 & 0 & 0 & 0 & 7 & \textbf{73} \\ 
\hline \hline
\end{tabular}
\end{table}

We finally examined the performance of the domain adaptation approach that was presented in Subsection III.C. 

We started by training the CNN-RNN network with only the MRI triplets in $\mathcal{P}'$ as inputs. The obtained performance when the trained network was applied to the test set $\mathcal{Q}'$ was only 70,6\%. For illustration of the proposed developments we extracted the $\textbf{r}'$ latent variables from this trained network and classified them to a set of  respectively extracted cluster centroids.      
Table~\ref{table:results5} presents the results of this classification task, which is consistent with the acquired DNN performance. It can be seen that the MRI information leads DNN prediction towards the patient class, which indeed contained more samples in the train dataset. Most errors were made in the non patient class (subjects 1 and 2).

\begin{table}[!t]
\caption{MRI-based Classification of 6 subjects' data in clusters $\textbf{t}_1$-$\textbf{t}_5$}
\label{table:results5}
\centering
\begin{tabular}{|c|c|c|c|c|c|}
\hline \hline
Test case & $\textbf{t}_1$ & $\textbf{t}_2$ & $\textbf{t}_3$ & $\textbf{t}_4$ & $\textbf{t}_5$    \\
\hline
Non Patient 1 & \textbf{181} & 74 & 179 & 8  & 0 \\ 
Non Patient 2 & 14 & 4 & \textbf{44}  & 33 & 5 \\ 
Patient 1 & 16 & 0  & \textbf{53} & 49 & 2 \\ 
Patient 2  & 6 & 0 & \textbf{83} & 80  & 15 \\ 
Patient 3 & 26 & 3 & \textbf{130} & 35 & 10 \\ 
Patient 4 & 12 & 0 & \textbf{51} & 11 & 6\\
\hline \hline
\end{tabular}
\setlength{\arrayrulewidth}{.4pt}

\end{table}

We then examined the ability of the proposed approach, to train the CNN-RNN network using the modified Loss Function, using various values of $\eta$; here we present the case when using a value equal to 0.5. 

The obtained performance when the trained network was applied to the test set $\mathcal{Q}'$ was raised to 81,1\%. For illustrating this improvement we also extracted the $\textbf{r}'$ latent variables from this trained network and classified them to one of the five annotated original cluster centroids $\mathcal{U}$. 

Table~\ref{table:results6} presents the results of this classification task. It is evident that minimization of the modified Loss Function managed to force the extracted latent variables get closer to cluster centroids which belonged in the correct class for Parkinson's diagnosis.

\begin{table}[!t]
\caption{MRI-based Classification of 6 subjects' data, after domain adaptation, in clusters $\textbf{t}_1$-$\textbf{t}_5$}
\label{table:results6}
\centering

\setlength{\arrayrulewidth}{.4pt}
\begin{tabular}{|c|c|c|c|c|c|}
\hline \hline
Test case & $\textbf{t}_1$ & $\textbf{t}_2$ & $\textbf{t}_3$ & $\textbf{t}_4$ & $\textbf{t}_5$    \\
\hline
Non Patient 1 & \textbf{176} & 147 & 114 & 5 & 0 \\ 
Non Patient 2 & 13 & \textbf{41} & 25 & 18 & 3  \\ 
Patient 1 & 13 & 0 & \textbf{70} & 35  & 2 \\ 
Patient 2 & 5 & 0 & \textbf{116} & 54 & 9\\ 
Patient 3 & 20 & 2  & \textbf{140} & 34 & 8 \\ 
Patient 4 & 9 & 0 & 31 & 5 & \textbf{35} \\
\hline \hline
\end{tabular}
\end{table}

\section{Conclusions and Future Work}
The paper proposed a new approach for extracting latent variables from trained DNNs, in particular CNN and CNN-RNN architectures, and using them in a clustering and nearest neighbor classification method for achieving high performance and transparency in Parkinson's diagnosis. We have used augmentation of the MRI and DaT Scan data in a recent Parkinson's database and provide the resulting datasets upon request from \url{mlearn.lincoln.ac.uk}. 

A DNN retraining procedure was presented, which is able to preserve the knowledge provided by annotated formerly extracted clustered latent variables. Moreover, a domain adaptation approach has been developed, which is able to use the extracted clustered latent variable information for improving the performance of the DNN architecture when presented with less input (only MRI) data.

An experimental study has been developed, using the above datasets, which illustrates the ability of the proposed approach to achieve high perfomance.

Future work will be based on a close collaboration of National Technical University of Athens and University of Lincoln with IBM, particularly relating the presented research to the IBM Watson Health initiative. The target will be generation of novel  performance-aware and transparent systems for better diagnosis of neurodegenerative diseases like Parkinson's, based on a combination of MRI and other images, epidemiological data, historical data of treatments and clinical data.

\section*{Acknowledgment}
\addcontentsline{toc}{section}{Acknowledgment}
The authors wish to thank the Department of Neurology of the Georgios Gennimatas General Hospital in Athens, Greece, and particularly Dr Georgios Tagaris, for the creation and provision of the main Parkinson's dataset and for his collaboration in the evaluation of the results of the performed analysis. 

%
%





\bibliographystyle{IEEEtran}
\bibliography{IEEEabrv,IEEEexample}

\vspace{12pt}

\end{document}